# Semantic Autoencoder and Its Potential Usage for Adversarial Attack

Yurui Ming, Cuihuan Du, and Chin-Teng Lin, *Fellow, IEEE*

*Abstract*—Autoencoder can give rise to an appropriate latent representation of the input data [1, 2], however, the representation which is solely based on the intrinsic property of the input data, is usually inferior to express some semantic information. A typical case is the potential incapability of forming a clear boundary upon clustering of these representations. By encoding the latent representation that not only depends on the content of the input data, but also the semantic of the input data, such as label information, we propose an enhanced autoencoder architecture named semantic autoencoder. Experiments of representation distribution via t-SNE shows a clear distinction between these two types of encoders and confirm the supremacy of the semantic one, whilst the decoded samples of these two types of autoencoders exhibit faint dissimilarity either objectively or subjectively. Based on this observation, we consider adversarial attacks to learning algorithms that rely on the latent representation obtained via autoencoders. It turns out that latent contents of adversarial samples constructed from semantic encoder with deliberate wrong label information exhibit different distribution compared with that of the original input data, while both of these samples manifest very marginal difference. This new way of attack set up by our work is worthy of attention due to the necessity to secure the widespread deep learning applications [3, 4].

## I. Introduction

AUTOENCODER finds various applications among tasks such as dimension reduction, data compression, etc., [1]. Usually, the more distinctive information contained in the latent representation, the more promising results achieved out of the applications [5]. However, these applications that depend on the quality of latent representation are recursively conditioning on the intrinsic characteristic of the input data. Although the latent representation might carry some semantic information of the input data, such as the class information, this is not always guaranteed.

Besides that, interpretation of the autoencoder and latent representation may benefit the understanding of other research, such as the working principles of visual system. Previously, it is thought that the higher-order areas of the visual cortices encode more global information by sacrificing some local fine details. In [6], research shows that neuronal clusters in area V4 still codes high-acuity information. However, considering that some structures of V1 demonstrate distinctive features, such as colour and orientation of the input images, it might be conjectured that retaining of high acuity information through the cortical visual hierarchy from V1 to V4 resemble the decoder which decodes the encoded information in the V1 area. Therefore, to continue study the autoencoder and to understand the correspondence between latent representation and decoded data might unveil more correlations between artificial neural networks and biological neural networks.

In this paper, we propose to enhance the contextual information in latent representation of autoencoders, especially from the semantic perspective. To do this, in contrast to the conventional approach, which just fine tunes the autoencoder based on the comparison between input data and decoded data, we simultaneously consider properties of the input data and the classes each data sample belonging to. Therefore, we parallelly chain a classifier to the encoder besides the decoder. In this way, the latent representation not only struggles to abstract information to better reconstruct the original data, but also distributes extra information such as class information to better categorize the input data, thus enforces certain semantic information buried into the latent representation. It is mentioned that although there has already been literatures wording "semantic" autoencoder [7], however, it actually adds a decoder to reverse the word embedding process to improve the semantic for such embeddings, hence, it is in essence a vanilla autoencoder, though it is termed semantic autoencoder.

The remaining parts of this paper are structured as follows. We firstly review the traditional autoencoder and recap the neural structure and concept. Then we depict the motivation and network structure of our proposed semantic autoencoder, which combines the autoencoder and a classifier. By experimenting on the commonly used MNIST dataset, we illustratively confirm the exclaim about lack of semantic information of the latent representation for vanilla autoencoder, contrasting with the proposed semantic autoencoder via t-SNE [8]. Thirdly, we consider learning algorithms based on autoencoders, to demonstrate the distribution shift by comparing the case of original input data and the adversarial counterparts constructed via the semantic autoencoder with deliberately wrong label information. Our experiment shows that indistinguishable image samples can render different latent distributions, which might exert impact on the performance of the original application. We believe continuing research on this subject might lead to new perspective of adversarial attacks and corresponding preventions to deep learning models.

## II. Autoencoder Review

Autoencoder is an unsupervised machine learning technique that utilizes neural networks to learn an effective coding of input data [1, 2]. The data can be labelled or unlabelled, though the label information is usually out of consideration in this circumstance. By exerting some constraints on the network structure, either a compressed or a sparse knowledge representation of the original input data could be learned [9]. Usually, this learned latent representation yields good performance of applications based on it. For example, feeding these intermediate latent codes to support vector machine (SVM) often bears higher accuracy than from other feature extractors [10]. Having said that, to investigate the meaning of the learned representation is not sufficiently carried out and there is still a margin between the application and theory.

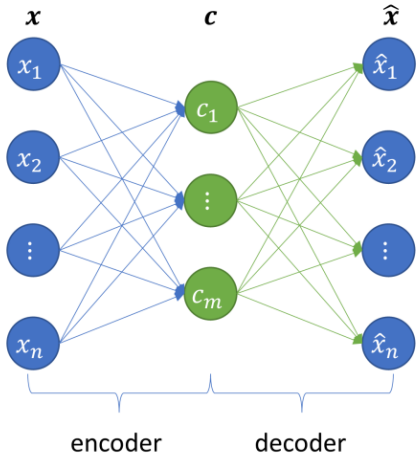

**Fig. 1.** An illustrative structure of conventional autoencoders.

The general architecture of the autoencoder is illustrated in Fig. 1. The sample $x$ in the input space is transformed into a code $c$ in the latent space by a network called encoder. To optimize the encoding process, we conceptually revert it as a decoding process to reconstruct $x$ by another network called decoder. Denote $c = T_{\text{enc}}(x)$ and $\hat{x} = T_{\text{dec}}(c)$, usually $T_{\text{enc}}$ is not invertible, which means $T_{\text{dec}} \circ T_{\text{enc}} \neq T_{\text{id}}$, here $T_{\text{id}}$ denotes the identity transformation. However, by minimizing $L(x, \hat{x})$, the divergence between the original data $x$ and reconstructed input $\hat{x} = T_{\text{dec}} \circ T_{\text{enc}}(x)$, it can enforce the encoder to capture the essential information from the input space.

## III. SEMANTIC AUTOENCODER

In the previous section, we show that the traditional autoencoder only learn by minimizing the difference between original data and reconstructed data, which means the network solely takes the content of input data into consideration without any other information. However, if the data are with label information, by incorporating this information into the learned latent representation, it might potentially benefit later applications. The reason is, intrinsic property of input data relies more on local textures, while class information is more conditioning on global features [11]. By simultaneously considering the content of the input data and the corresponding class information, the components of encoded representation are expected to be more orthogonal in the latent space. Usually, the orthogonality can confer a robust decision boundary [12, 13], and this will reciprocally enforce the encoder to learn features in orthogonal forms. Put it together, this way can benefit applications which use the latent representation as intermediate result such as in [10].

The above understanding motivates our proposal for a semantic autoencoder as in Fig. 2. The overall architecture is quite straightforward. Taking image data encoding for example, we add an extra classifier to enforce the learned representation beneficial for image classification besides the conventional decoder. In this way, we consequently embed some semantic information, which is about the image categories, into the latent representation.

To demonstrate the supremacy of our proposed model, we use the MNIST dataset with an instance configuration of the general network structure in Fig. 2. The layers and parameters of the network is shown in TABLE I. We train both the vanilla autoencoder and the semantic autoencoder with the same hyperparameters and illustrate the corresponding results. In detail, using a batch size of 32 and initial learning rate of 0.001, we training the network for 10,000 iterations by Adam optimizer. To prevent potential overfitting, learning rate is decayed by a factor of 0.96 for every 100 iterations. Then we load the trained weights to apply to the test set. We repeat such a process for the vanilla autoencoder (without a classifier) and the semantic one (with the classifier) respectively and compare their results.

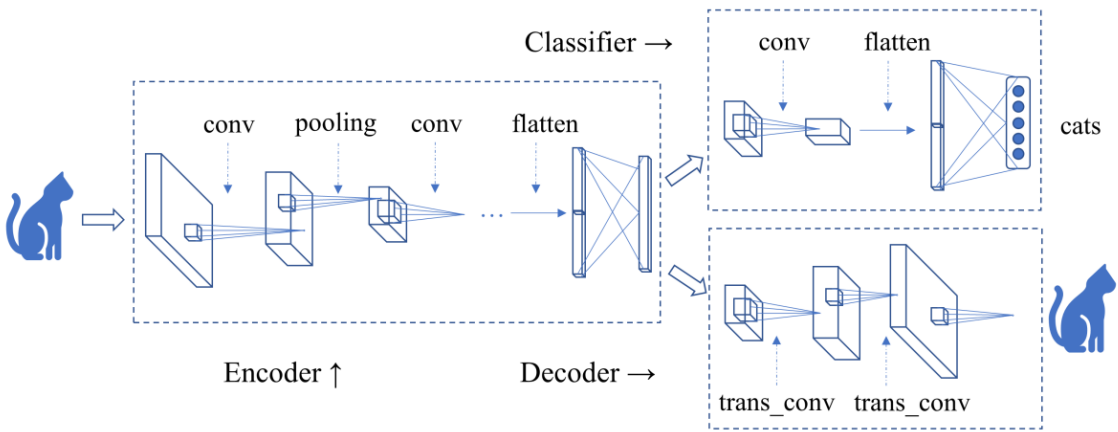

**Fig. 2.** The overall network architecture of the proposed semantic autoencoder.

TABLE I
NETWORK ARCHITECTURE CONFIGURATION

| Model | Layer ID | Layer Type | #Filters | Size | Stride | Act. | Pad. | Comment |
|---|---|---|---|---|---|---|---|---|
| Encoder | 1 | Conv2D | 32 | (5, 5) | 1 | ReLU | Same | Shared both by vanilla autoencoder And semantic autoencoder |
| | 2 | MaxPool2D | - | (2, 2) | 2 | - | - | |
| | 3 | Conv2D | 64 | (5, 5) | 1 | ReLU | Same | |
| | 4 | MaxPool2D | - | (2, 2) | 2 | - | - | |
| | 5 | BatchFlatten | - | - | - | - | - | |
| | 6 | Linear | - | 49*4 | - | sigmoid | - | |
| Decoder | 1 | BatchReshape | - | - | - | - | - | Shared both by vanilla autoencoder and semantic autoencoder |
| | 2 | Conv2DTranspose | 64 | (5, 5) | 2 | ReLU | Same | |
| | 3 | Conv2DTranspose | 32 | (5, 5) | 2 | ReLU | Same | |
| | 4 | Conv2D | 1 | (1, 1) | 1 | sigmoid | - | |
| Classifier | 1 | BatchReshape | - | - | - | - | - | Only present in the semantic autoencoder |
| | 2 | Conv2D | 32 | (3, 3) | 1 | ReLU | SAME | |
| | 3 | BatchFlatten | - | 1 | - | - | - | |
| | 4 | Linear | - | 256 | - | ReLU | - | |
| | 5 | Linear | - | #classes | - | - | - | |

First, we show the exemplified reconstructed hand-written digits in Fig. 3. To measure the quality, two metrics are employed by comparing these decoded images with the original ones. The first is the peak signal-to-noise ratio (PSNR), which is more numeric and objective oriented, with less concerns from the subtlety of the human visual system. The second is the structural similarity (SSIM), which is a perception-based measurement to quantify the change of altered or degraded images from the original ones, especially from the perceptual perspective. We calculate these indices of each reconstructed image in Fig. 3 for the vanilla case and semantic case respectively, and show these statistics in TABLE II.

From TABLE II, it indicates incorporation of extra categorical information to the latent representation only exerts marginal effect on the quality of restructured data. Although there is a slight decrease of SSIM, which is also perceivable if by comparing Fig. 3(B) and Fig. 3(C), however, it is overall subjectively acceptable.

The next step is trying to demonstrate the characteristics of the latent representations from different autoencoders by t-SNE. As showed in Fig. 4, it is manifest that latent representation endowed with the class information has more packed individual clusters than the vanilla case. Since the class information can distinguish samples from each other, once injected into encoded data during optimization via gradient descent, it can be expected that lower intra-class variance and higher inter-class variance are exhibited, which strongly support our claim that semantic representation is rich in information and can potentially benefit down-stream applications. For example, decision boundary is much easier drawn in Fig. 4(B) than in Fig. 4(A).

TABLE II
QUALITY METRICS OF RECONSTRUCTED IMAGES

| | Average PNSR | Average SSIM |
|---|---|---|
| Vanilla Autoencoder | 24.89 | 96.39% |
| Semantic Autoencoder | 25.10 | 95.82% |

IV. ADVERSARIAL ATTACK

We investigate the potential adversarial attack in the scenario that learning algorithms are based on autoencoders [5, 10]. These applications rely on the quality of latent representation achieved via the autoencoder. Previous section demonstrates that embedding of semantic information can alter the distribution of latent representation. Therefore, to deliberately incorporate wrong class information into latent representation is expected to alter the latent distribution and sever the performance. This part outlines such an attack by employing our proposed semantic autoencoder.

We begin by deducing from the theoretic perspective to reveal the potentiality of such attacks. First, given input data *I*

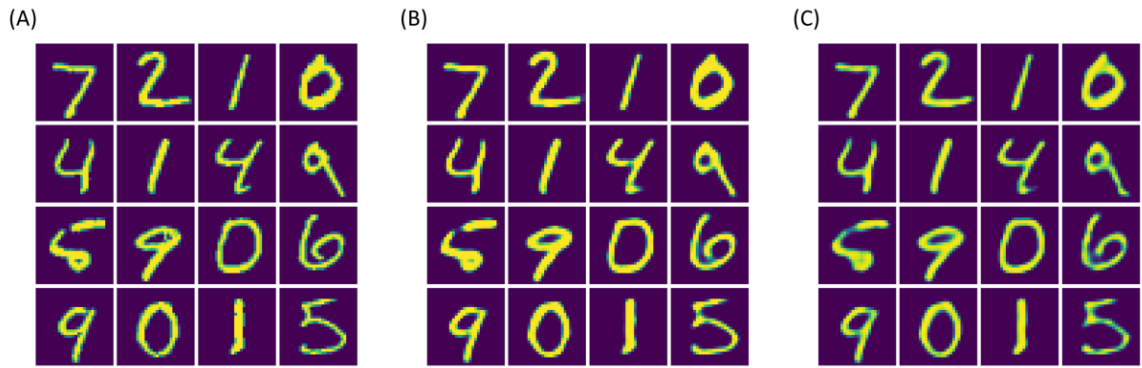

**Fig. 3.** Hand-written digits from the experiment: (A) Original ones; (B) Decoded ones from the vanilla autoencoder; (3) Decoded ones from the semantic autoencoder.

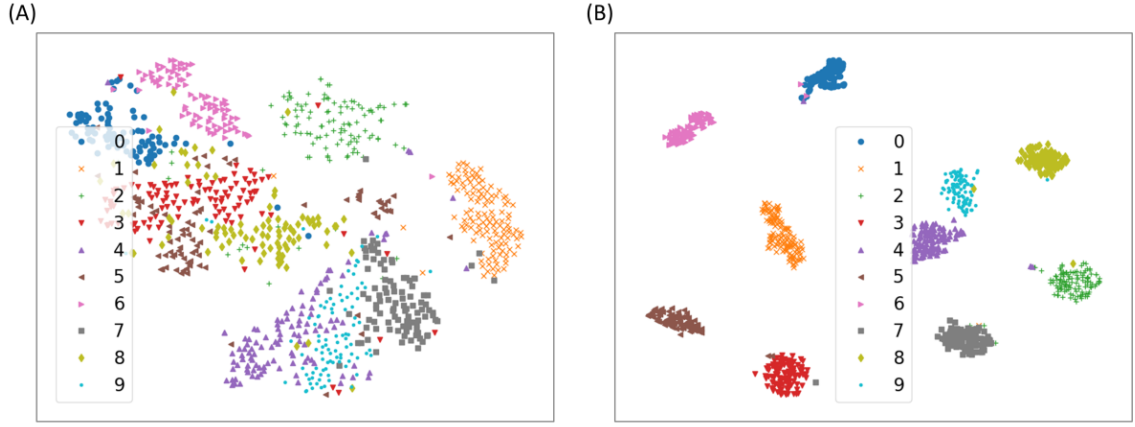

**Fig. 4.** Illustration of latent representation via t-SNE. (A) The vanilla case; (B) The semantic case.

and considering the vanilla case, let the encoder denoted by a function $e_v$, the corresponding decoder as $d$, the latent representation as $L_v$, and the decoded data by $I_v$. Then we have the following formulas:

$$L_v = e_v(I) \qquad (1)$$

$$I_v = d(L_v) \qquad (2)$$

Since we optimize the autoencoder by having $\|I - I_v\| \to 0$, let $\mathrm{id}_X$ denote general identical mapping, then we assume the following (3) holds:

$$e_v \circ d = d \circ e_v = \mathrm{id}_X \qquad (3)$$

Note in (3) the equal sign only means equivalence in an approximation sense but not exactly equal to.

Then we consider the introduction of adversarial attack by connecting the classifier to the system for embedding some adversarial information. We notice from the last section that correct label information enforces the coherence of intra-class samples and diversity of inter-class samples, therefore, random shuffle of the labels during training of the semantic autoencoder might induce an opposite effect, that is making the overall samples more indistinguishable. Let the encoder involved in semantic case denoted by $e_s$ and we reuse the same decoder as in the vanilla case (which means the semantic autoencoder shares the trained weights of the vanilla decoder), we have the following (4) and (5):

$$L_s = e_s(I) \qquad (4)$$

$$I_s = d(L_s) \qquad (5)$$

Now consider an input sample $I_a$ with values defined in (6):

$$I_a = I + I_s - I_v \qquad (6)$$

Since $\|I_s - I_v\| = \|I_s - I - (I_v - I)\| \le \|I_s - I\| + \|(I_v - I)\|$, and we know that during optimisation, both $\|I_s - I\|$ and $\|(I_v - I)\|$ approach to zero, therefore, $\|I_a - I\| = \|I_s - I_v\| \to 0$, which means $I_a$ manifests minimal difference from $I$ in theory.

Further, if we assume $e_v$ and $d$ retain some linearity, then it can be reached that $e_v(I_a) = e_v(I) + e_v(I_s - I_v) = L_v + e_v(d(L_s) - d(L_v)) = L_v + e_v \circ d(L_s - L_v) = L_v + e_v \circ d(L_s - L_v) = L_v + \mathrm{id}_X(L_s - L_v) = L_v + L_s - L_v = L_s$. That is:

$$e_v(I_a) = L_s \qquad (7)$$

Because during training of the semantic autoencoder, we deliberate embed some false information into the latent representation $L_s$, which means if we substitute the original image data $I$ with $I_a$, the data retain minimal change but the corresponding latent representations might exhibit obvious distributive disparities.

Based on the above deduction, we choose the commonly-used CIFAR-10 dataset [14] to exemplify such an attack. We instantiate a network structure guided by Fig. 2, and illustrate the corresponding t-SNE illustration of both cases. However, since during our experiment, the original distributions of samples from all categories fail to render clear boundaries between them, it is hard to speculate the adversarial effect if we simultaneously illustrate all the categories. To mitigate this, we only choose the first two classes, namely, airplane and automobile, and experimenting with samples from these two categories.

The overall architecture and training process resemble the test using MNIST dataset in the last section. Because this information can be directly referenced from our open-sourced code, we omit the technical details here. We only point out that we process each channel of the input image separately then combine them together. Besides, to construct adversarial attack, which means to utilize some faulty information, here during the training of the semantic autoencoder, we randomly shuffle the labels before calculating the classification loss, which consequently gets the incorrect label information injected into the latent representation via the gradient descent process.

After training of each autoencoder, we construct the adversarial samples guided by (6). We first compare some representative original samples and the corresponding adversarial ones. They are illustrated in Fig. 5 with image degradation quality metric PNSR as 39.93 and SSIM as 96.55% in average, which are promising values indicating marginal difference between the original images and the adversarial ones. These statistics indicate neglectable difference between the original data and adversarial ones even under scrutiny.

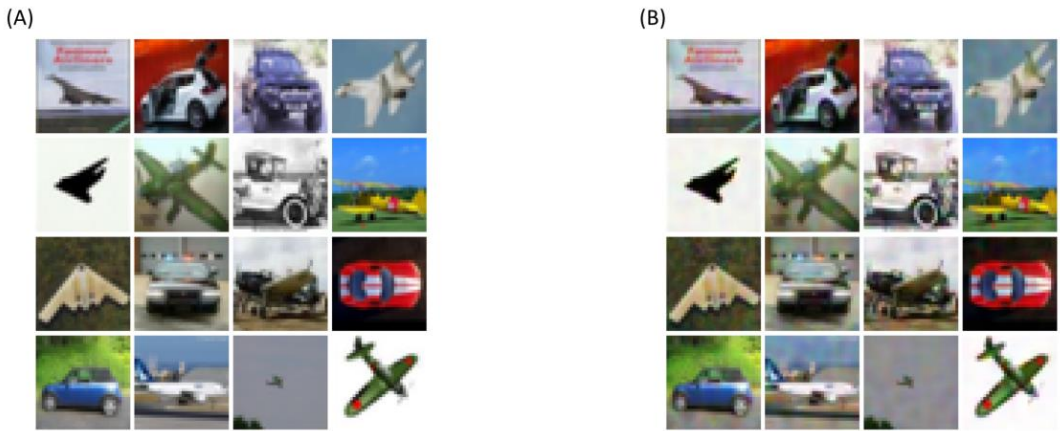

**Fig. 5.** Sample illustration. (A) Original images; (B) Adversarial images.

Then we illustrate the representation distributions in each case using t-SNE in Fig. 6. It can be observed from Fig. 6(A) that the latent representations from original image data exhibits certain polar distribution from the inter-class perspective. For intra-class situation, the distribution is more compact. However, in Fig. 6(B), the latent representations encoded from the adversarial image data by the same encoder are more dispersed. Such a change of conformality with the distribution of original data if is deliberately introduced, might lead to misunderstanding or even false information towards the problem to be solved, especially the properties of samples are manually screened via some visualization tools.

## V. DISCUSSION

To reach (7) we assume that assume $e_v$ and $d$ retain some linearity. This can be roughly satisfied if all the inputs and outputs of the intermediate layers are around 0 if some non-linear activation function are used, because commonly-used activation functions such as sigmoid and tanh are approximately linear near 0. Therefore, even if it is still coarse, we normalize the image data into the range [0, 1]. We assume that ReLU is also a candidate since at least it is of some semi-linear property. We are still working on more suitable conditions to have our deduction more rigorous.

Further, our work is still in a preliminary stage and is more like a proof of concept. The current work just demonstrates working collectively with many samples to change the population distribution in an adversarial way, and we are still conceiving how to construct an effective adversarial attack using ideas from our paper and planed as a future work.

## V. CONCLUSION

This paper investigated incorporation of extra information, such class information into the latent representation obtained from an encoder to construct an enhance architecture called semantic autoencoder. We demonstrated via experiment with MNIST dataset that embedding label information can enforce the samples of intra-class distributed in a more compact way, while samples of inter-class exhibit more obvious discrepancy. We also deduced the procedures to induce adversarial attack via embedding wrong information based on some assumptions. The result of experiment using CIFAR-10 dataset showed adversarial examples obtained from semantic autoencoder with randomized shuffled label information can cause the distribution of adversarial latent representation obvious shifting from the original distribution. We also discussed the aspects to improve our work in future.

## REFERENCES

[1] M. A. Kramer, "Nonlinear principal component analysis using autoassociative neural networks," *Aiche Journal,* vol. 37, pp. 233-243, 1991.

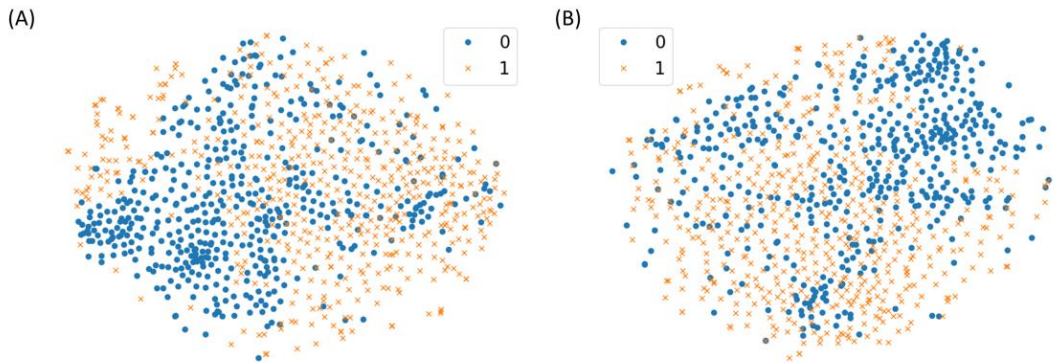

**Fig. 6.** Illustration of latent representation via t-SNE, with 0 for airplane and 1 for automobile. (A) The normal case; (B) The adversarial case.